\documentclass[final,1p,times]{elsarticle}
\usepackage{multirow}
\usepackage{booktabs}

\usepackage{graphics}
\usepackage{graphicx}
\usepackage{epsfig}
\usepackage{subfigure}

\usepackage{graphicx}
\usepackage{enumerate}
\usepackage{amssymb}
\usepackage{amsmath}
\usepackage{CJK}

\usepackage{algorithm}
\usepackage{algpseudocode}
\usepackage{makecell}
\usepackage{latexsym}

\usepackage{url}
\usepackage{xcolor}
\usepackage{xspace}
\usepackage{comment}

\newcommand{\eg}{\emph{e.g.,}\xspace}
\newcommand{\ie}{\emph{i.e.,}\xspace}

\newcommand{\baby}{W\textsc{sptm}\xspace}

\begin{document}

\begin{frontmatter}



\title{Weakly Supervised Prototype Topic Model with Discriminative Seed Words: Modifying the Category Prior by Self-exploring Supervised Signals}


\author[1,2]{Bing Wang}
\author[1,2]{Yue Wang}
\author[1,2]{Ximing Li}
\author[1,2]{Jihong Ouyang}

\address[1]{College of Computer Science and Technology, Jilin University, China\fnref{label3}\\
         }
\address[2]{Key Laboratory of Symbolic Computation and Knowledge Engineering of Ministry of Education, Jilin University, China\fnref{label4}}

\begin{abstract}
Dataless text classification, \ie a new paradigm of weakly supervised learning, refers to the task of learning with unlabeled documents and a few predefined representative words of categories, known as \textit{seed words}. The recent generative dataless methods construct document-specific category priors by using seed word occurrences only, however, such category priors often contain very limited and even noisy supervised signals. To remedy this problem, in this paper we propose a novel formulation of category prior. First, for each document, we consider its label membership degree by not only counting seed word occurrences, but also using a novel \textit{prototype scheme}, which captures pseudo-nearest neighboring categories. Second, for each label, we consider its frequency prior knowledge of the corpus, which is also a discriminative knowledge for classification. By incorporating the proposed category prior into the previous generative dataless method, we suggest a novel generative dataless method, namely Weakly Supervised Prototype Topic Model (\baby). The experimental results on real-world datasets demonstrate that \baby outperforms the existing baseline methods.

\end{abstract}


\end{frontmatter}



\section{Introduction}
\label{1}

Automatic text classification is a well-established yet still challenging research topic in the information retrieval and machine learning communities. Generally speaking, traditional text classifiers are mainly developed in the paradigm of supervised learning, which always requires big training corpora of massive labeled documents to ensure high performance. However, manually collecting labeled documents is extremely expensive, especially for big real-world applications that require millions of labeled documents.

Recently, dataless text classification, \ie a new paradigm of weakly supervised learning, has attracted increasing attention from the community \cite{AAAI2004,AAAI2008,NIPS2008,SIGIR2008,SIGIR2013b,SIGIR2014,AAAI2015,STM2016,coling2018,CIKM2018,Mengyu2018,Chenliang2019,KAIS2019,Mengyu2020}. The target is to build text classifiers by training over unlabeled documents with predefined representative words of categories (called \textbf{seed words}), instead of labeled documents. Because manually selecting a small set of seed words for a given dataset is much cheaper than labeling all documents \cite{SIGIR2008}, the dataless method can effectively save many human efforts and become a promising supplement to traditional supervised text classifiers.

\vspace{3pt}
\noindent{\textbf{Prior arts:}}
To our knowledge, the existing dataless methods are mainly divided into two categories: {\textbf{discriminative dataless methods}} \cite{AAAI2004,AAAI2008,SIGIR2008,NIPS2008,coling2018} and {\textbf{generative dataless methods}} \cite{SIGIR2013b,SIGIR2014,AAAI2015,STM2016,CIKM2018}.

The discriminative dataless methods mainly extend traditional discriminative methods to train classifiers over unlabeled corpora with only seed words. An early method proposed in \cite{AAAI2004} manually selects seed words by referring to their information gain values, which are computed on the k-means clustering results over unlabeled documents. The method employs the selected seed words to construct a pseudo training dataset, and then iteratively train a naive Bayes classifier on it using semi-supervised expectation maximization. A recent method, named pseudo-label based dataless naive Bayes (PL-DNB) \cite{coling2018}, simultaneously trains the naive Bayes classifier and updates the pseudo training dataset using both the seed word information and temporary prediction results. Additionally, the maximum entropy-based dataless classifier \cite{SIGIR2008} supposes that a document is more likely to belong to the categories, whose seed words have occurred in this document. It builds an objective of the distance between the expected label distributions of documents containing seed words and the corresponding reference label distributions of seed words. Another interesting work \cite{AAAI2008} trains a dataless classifier using Wikipedia concepts.

The generative dataless methods focus on incorporating the supervision provided by seed words into the generative assumption of corpora. The representatives are mainly based on topic modeling such as latent Dirichlet allocation (LDA) \cite{Blei2003}, which are widely applied in various tasks \cite{Blei2012,Yang2014,Fu2018}. The ClassifyLDA \cite{SIGIR2013b} and its extension TLC++ \cite{SIGIR2014} manually make correspondences between labels and temporary topics, and then continue to infer the model using those topics as initialization. Recently, seed-guided topic model (STM) \cite{STM2016} and Laplacian seed word topic model (LapSWTM) \cite{CIKM2018}, incorporate the supervision information by constructing document-specific category priors with seed word occurrences, and further schemes, \eg manifold regularizer \cite{CIKM2018}, are used to spread the supervision.

\begin{table}
\renewcommand\arraystretch{1.5}
\caption{Examples of seed words (\ie label descriptions here) and statistics of seed word occurrences across \emph{Reuters} and \emph{Newsgroup}. NonSW: the number of documents that contain no seed words. TrueMark: the number of documents that contain the seed words from relevant labels.}
\label{Problem}
\centering
\small
\begin{tabular}{m{38pt}<{\centering} | m{134pt}<{\centering} | m{80pt}<{\centering} | m{82pt}<{\centering}}
\Xhline{1.2pt}
\textbf{Dataset}   &  \textbf{Label ID (Description)}  & \textbf{NonSW/Total (\%)}  & \textbf{TrueMark/Total (\%)} \\
\Xhline{1pt}
\textit{Reuters}        & 1 (acquisition)\:\: 2 (coffee)\:\: 3 (crude)\:\: $\cdots$ \:\: 9 (sugar)\:\: 10 (trade)   & 4,512 / 7,285 (62\%)  & 1,877 / 7,285 (26\%) \\
\hline
\textit{Newsgroup}      & 1 (atheism)\:\: 2 (computer graphics)\:\: 3 (computer os microsoft windows)\:\: $\cdots$ \:\: 19 (politics)\:\: 20 (religion)   & 9,941 / 18,846 (52\%)  & 5,773 / 18,846 (30\%) \\
\Xhline{1.2pt}
\end{tabular}
\end{table}

\vspace{3pt}
\noindent{\textbf{Our motivation and contributions:}}
As reported in \cite{AAAI2015,STM2016,CIKM2018}, the generative dataless methods have empirically achieved very competitive performance. For example, the recent LapSWTM method has significantly outperformed other existing dataless methods on the commonly used benchmark datasets. Unfortunately, a tough problem remains. In the generative dataless family, a key step is to construct document-specific category priors to incorporate the supervision information. However, the category priors are computed by only counting seed word occurrences, resulting in very limited and even noisy supervised signals, especially when the seed words are scarce. \textit{First}, many training documents may contain no seed words, so that their corresponding category priors involve supervision information by no means. \textit{Second}, many training documents may contain or contain only seed words of irrelevant categories, resulting in noisy supervision. To visualize the two problems, we show statistics of seed word occurrences on two benchmark datasets given seed words extracted from the label description in Table \ref{Problem}. For example, in terms of \emph{Newsgroup}, about \underline{52\%} documents have no seed words and only about \underline{30\%} documents contain seed words from relevant categories.

In this paper, we aim to modify the category prior by self-exploring the available data with seed words, so as to capture more accurate supervised signals. To this end, we propose a novel formulation of category prior, which consists of two components. (1) The first component describes the label membership degree prior for each document. Naturally, we consider it by not only counting seed word occurrences, but also using a novel \textbf{prototype scheme}. In this scheme, each category is represented by a prototype vector computed using word co-occurrences with seed words, and each document is considered to be associated with its pseudo-nearest neighboring categories measured by those prototype vectors, hence enriching supervised signals. (2) The second component describes the label frequency prior at the corpus level, which is also a discriminative knowledge in classification. For each label, we estimate its label frequency using seed word occurrences in the corpus, following the assumption that the category with more seed word occurrences tends to be a larger category and is more likely to appear in the test corpora, and vice versa. \textit{In summary}, we obtain the novel category prior by combining the two components. We then incorporate this category prior into the prior method LapSWTM, leading to a novel dataless model, namely \textbf{W}eakly \textbf{S}upervised \textbf{P}rototype \textbf{T}opic \textbf{M}odel (\textbf{\baby}). Additionally, since in the context of dataless text classification with seed words, the supervision is explored at the word level, the model may be sensitive to less discriminative words, \eg domain-specific stopwords. To solve this, we employ a term weighting method named Log weight to punish less discriminative words. To evaluate \baby, we compare it against both existing dataless methods and supervised methods. The empirical results on commonly used benchmark datasets indicate that \baby outperforms the dataless baseline methods, and it is even on a par with traditional supervised methods in some settings.

The major contributions of this paper are summarized as follows:
\begin{itemize}
    \item [1] We propose a novel formulation of category prior that can enrich and modify the supervision information provided by seed words.
    \item [2] We incorporate the novel category prior into the prior method LapSWTM, leading to \textbf{\baby}.
    \item [3] We apply the term weighting to further improve the performance.
    \item [4] The empirical results demonstrate that the effectiveness of\baby.
\end{itemize}

The rest of this paper is organized as follows: In Section 2, we review some related works on dataless text classification. In Section 3, we briefly review the prior method LapSWTM. We describe the proposed \baby method in Section 4. In Sections 5 and 6, the experimental results and conclusions are presented.

\section{Related Work}
\label{5}

In this section, we mainly review the most related works on generative dataless methods \cite{SIGIR2013b,SIGIR2014,AAAI2015,STM2016,CIKM2018}. An early model ClassifyLDA \cite{SIGIR2013b} is built on a three-stage learning procedure. First, it runs the standard LDA over unlabeled documents to obtain a set of topics; second, an annotator manually assigns each label to one of those topics; finally, it aggregates the topics associated with the same label as a single one, and then restarts the LDA inference using those aggregated topics as initializations. The topic label classification (TLC++) model \cite{SIGIR2014} modifies ClassifyLDA by allowing to assign multiple labels to each topic (referring to stage 2 in ClassifyLDA). However, those methods rely on the quality of annotation assignment, which may be some bias. Most recently, two dataless models with seed words, \ie STM \cite{STM2016} and LapSWTM \cite{CIKM2018}, have been proposed. Both of them directly make a one-to-one correspondence between labels and category topics, and incorporate the supervision information by constructing document-specific category priors that are based on seed word occurrences. STM and LapSWTM are built on the assumptions of Dirichlet multinomial mixture and LDA, respectively. As reported in \cite{STM2016,CIKM2018}, they, especially LapSWTM, consistently outperform other existing dataless methods. However, both of them still suffer from a tough problem, where the document-specific category prior may be inaccurate with only seed word occurrences. In contrast to them, our \baby uses the prototype scheme to better capture the label membership degree prior and further considers the label frequency prior, leading to more accurate category priors. Empirical results shown in the Section of Experiment indicate the superior performance of \baby.


Additionally, there are many previous supervised topic models that directly incorporate the supervision information into the unsupervised versions, including both works on single-label learning \cite{sLDA2007,DiscLDA2008,MedLDA2012,GibbsMedLDA2014,Yang2015,DTM2016,sLDAC2017,IPM2018_1,IPM2019} and multi-label classification \cite{ICDM2008,LLDA2009,PLDA2011,DLDA2012,DPMRM2012,Li2015,IPM2018_2,Chenliang2019,IPM2019_1}. These models have empirically achieved very competitive classification performance, however, they require labeled documents as inputs. In contrast, our \baby trains a classifier only using the much cheaper seed words, hence saving many human efforts on collecting labeled training corpora.

\section{Preliminary}
\label{2}

We briefly review the prior method LapSWTM \cite{CIKM2018}. For convenience, some important notations are outlined in Table \ref{Notations}. 


\vspace{3pt}
\noindent{\textbf{Model structure.}}
Given a corpus of $D$ documents, LapSWTM posits $K$ category topics and $G$ background topics. The category topic has a one-to-one correspondence with the category label, used to described category-specific semantics, while the background topic is used to describe the common semantics of corpora. Both kinds of topics, denoted by $\phi$ and $\widehat \phi$, are multinomial distributions over words, drawn from Dirichlet priors $\beta$ and $\widehat \beta$, respectively. Each document is simultaneously associated with multinomial distributions over category topics and background topics, denoted by $\theta$ and $\widehat \theta$, respectively. For each document \emph{d}, LapSWTM draws $\widehat \theta_d$ from a Dirichlet prior $\widehat \alpha$, and specifically draws $\theta_d$ from a \textbf{supervised Dirichlet prior} $\alpha_d$, constructed with seed words. For each word, a topic type indicator $c_{dn}$ will be first drawn from a Bernoulli distribution $\delta_{dn}$, and then (1) in the case of $c_{dn}=1$ the model draws a category topic $z_{dn}$ from $\theta_d$, and then $w_{dn}$ from $\phi_{z_{dn}}$; (2) in the case of $c_{dn}=0$, draws $w_{dn}$ from $\widehat \theta$ and $\widehat \phi$ similarly. 

\vspace{2ex}
\noindent{\textbf{Supervised Dirichlet prior $\alpha$.}}
The $\alpha$ serves as the supervised prior of category topics per document. Given a pre-defined set of seed words, it can be computed by using seed word occurrences, describing the label membership degree prior per document:
\begin{equation}
\label{Eq1}
\alpha_{dk} = \eta \frac{DF_{dk}}{\sum \nolimits_{i=1}^K DF_{di}} + \alpha_0, \quad \quad d=1,\cdots,D, \:\: k = 1, \cdots, K,
\end{equation}
where $DF_{dk}$ denotes the number of times that seed words of category \emph{k} appear in document \emph{d}; $\eta$ is the concentration parameter; and $\alpha_0$ is the smoothing parameter.

\vspace{2ex}
\noindent{\textbf{Bernoulli parameter $\delta_{dn}$.}}
In the model, the variable $\delta_{dn}$ is computed by the dot product $\theta_d^T \gamma_{w_{dn}}$, where each component $\gamma_{vk}$ describes the relevance degree between word \emph{v} and category topic \emph{k}, computed as follows:
\begin{eqnarray}
\!\!\!\!\!\!\!\!\!\!\!\!\!\!\!\!\!\! & & u_{vk} = \max\left(\frac{SC_{vk}}{\sum \nolimits _{i=1}^K SC_{vi}} - \frac{1}{K},0\right), \quad u_{vk} \leftarrow \frac{u_{vk}}{\sum \nolimits _{i=1}^V u_{ik}}, \nonumber \\
\!\!\!\!\!\!\!\!\!\!\!\!\!\!\!\!\!\! & &  \gamma_{vk} = \max\left(\frac{u_{vk}}{\sum \nolimits _{i=1}^K u_{vi}},\epsilon\right), \quad \quad \quad  k=1,\cdots,K,\:\:v=1,\cdots,V,\nonumber
\end{eqnarray}
where $V$ is the vocabulary size; $SC_{vk}$ is the number of co-occurrences between word \emph{v} and seed words of category \emph{k}; and $\epsilon$ is a smoothing parameter, empirically set to 0.01 by referring to \cite{STM2016,CIKM2018}.

\begin{table}[t]
\renewcommand\arraystretch{1.05}
\small
\caption{A summary of notations}
\label{Notations}
\begin{center}
\begin{tabular}{p{0.09\columnwidth}<{\centering}|p{0.80\columnwidth}}
\Xhline{1.2pt}
Notation & Description\\
\Xhline{1.2pt}
\emph{D}    &   number of documents \\
\emph{V}    &   number of words \\
\emph{K}    &   number of category topics, \ie category labels \\
\emph{G}    &   number of background topics \\
\Xhline{1.2pt}
$\phi$      &   category topic distribution over words\\
$\widehat \phi$      &   background topic distribution over words\\
$\theta$    &   document-specific distribution over category topics\\
$\widehat \theta$    &   document-specific distribution over background topics\\
$\delta$      &  Bernoulli distribution for choosing topic type \\
\Xhline{1.2pt}
$\beta$     &   Dirichlet prior of $\phi$ \\
$\widehat \beta$     &   Dirichlet prior of $\widehat \phi$ \\
$\alpha$    &   supervised Dirichlet prior of $\theta$ \\
$\widehat \alpha$   &   Dirichlet prior of $\widehat \theta$ \\
\Xhline{1.2pt}
$\alpha'$    &   supervised Dirichlet prior of $\theta$ in \textbf{\baby} \\
$\rho$      &  tuning parameter of $\alpha'$\\
$\eta$      &  concentration parameter of $\alpha'$\\
$\alpha_0$  &  smoothing parameter of $\alpha'$ \\
\emph{P}    &  number of pseudo-nearest neighboring categories in the \textbf{prototype scheme}\\
$\tau$      &  tuning parameter in the document-specific label membership degree prior\\
\Xhline{1.2pt}
\end{tabular}
\end{center}
\end{table}


\vspace{3pt}
\noindent{\textbf{Objective of LapSWTM.}}
LapSWTM incorporates a manifold regularizer $\mathcal{R}(\theta)$ to spread supervised signals through neighboring documents, defined below:
\begin{equation}
\mathcal{R}(\theta) = \frac{1}{2} \sum \limits _{k=1}^K \sum \limits_{i,j=1}^D \left(\theta_{ik} - \theta_{jk}\right)^2 W_{ij},
\end{equation}
The notation $W$ denotes the neighboring weight between document pairs, computed by:
\begin{eqnarray}
\quad \quad \quad {W_{ij}} = \left\{ \begin{array}{l}
 1 \quad {\rm{if}}\:\:\: d_i \in \Pi(d_j) \:\: {\rm{or}} \:\: d_j \in \Pi(d_i)\\
 0 \quad {\rm{otherwise}}\\
 \end{array} \right., \nonumber
\end{eqnarray}
where $\Pi(d)$ is the nearest neighbor set\footnote{Following \cite{CIKM2018}, we retain the top-5 neighbors for each document in $\Pi(d)$.} of document \emph{d}.

By incorporating $\mathcal{R}(\theta)$ into the log-likelihood function of the model given a training data $\mathbb{W}$, LapSWTM can learn the latent variables of interest, \ie $\{\theta,\phi,\widehat \theta, \widehat \phi\}$, by maximizing the following objective:
\begin{equation}
\label{Eq3}
\mathcal{L}(\theta,\widehat \theta,\phi,\widehat \phi) = \log p\left(\mathbb{W},\theta,\widehat \theta,\phi,\widehat \phi|\alpha,\widehat \alpha,\beta,\widehat \beta,\gamma\right) - \lambda \mathcal{R}(\theta),
\end{equation}
where $\lambda \ge 0$ is the regularization parameter

\section{The Proposed Model}
\label{3}

In this section, we first discuss the problem in LapSWTM and then introduce the proposed model, namely \textbf{W}eakly \textbf{S}upervised \textbf{P}rototype \textbf{T}opic \textbf{M}odel (\textbf{\baby}).

\subsection{Problem Description}
\label{3.1}


Revisiting the model structure of LapSWTM, we can observe that the supervision information is solely incorporated by the supervised Dirichlet prior $\alpha$, which describes the label, \ie category topic, membership degree prior of documents. However, since $\alpha$ is built on seed word occurrences, \textbf{it may involve less supervision information or introduce noisy supervision, especially for the situations with scarce seed words}. To be specific, by referring to Eq.\ref{Eq1}, for any document \emph{d} if it contains no seed words, resulting in the symmetric prior $\alpha_d$ without any supervision; and if it contains many seed words from irrelevant categories, resulting in a low-quality prior $\alpha_d$ with noisy supervision. Unfortunately, such problems often arise in many real-world applications. To visualize this, we illustrate several example statistics\footnote{Details of datasets and seed word sets are shown in Section 5.1} of \emph{Reuters} and \emph{Newsgroup} with two seed word sets $S^L$ and $S^D$ in Table \ref{Example}.\footnote{Kindly reminding that, the statistics of seed word occurrences (when $P=0$) with $S^L$ (\ie label descriptions) are exactly the statistics that have already been shown in Table \ref{Problem}.} Taking \emph{Newsgroup} with $S^L$ as an example, about half of documents contain no seed words, and only about a third of documents contain the seed words from relevant categories.

\subsection{\baby}
\label{3.2}

\begin{table}
\renewcommand\arraystretch{2}
\caption{Examples of statistics of marked labels on \emph{Reuters} and \emph{Newsgroup}. $S^L$ and $S^D$ are two different seed word sets. $P=0$ actually implies that the prototype scheme is not applied. NonSW: the number of documents that contain marked labels. TrueMark: the number of documents, whose relevant labels belong to marked labels.}
\label{Example}
\centering
\footnotesize
\begin{tabular}{m{22pt}<{\centering}|m{20pt}<{\centering}|m{64pt}<{\centering}|m{66pt}<{\centering}|m{66pt}<{\centering}|m{70pt}<{\centering}}
\Xhline{1.5pt}

\multirow{2}{*}{} & \multirow{2}{*}{$P$}  & \multicolumn{2}{c|}{\textit{Reuters}} & \multicolumn{2}{c}{\textit{Newsgroup}}    \\
\cline{3-6}
 & & NonSW/Total (\%)   & TrueMark/Total (\%)   & NonSW/Total (\%)    & TrueMark/Total (\%)    \\
\Xhline{1pt}
\hline
\hline

\multirow{2}{*}{$S^L$} & $P = 0$ & 4,512/7,285 (62\%) & 1,877/7,285 (26\%) & 9,941/18,846 (52\%) & 5,773/18,846 (30\%) \\
\cline{2-6}
 & $P = 1$ & 0/7,285 (0\%) & 5,012/7,285 (69\%) & 0/18,846 (0\%) & 12,712/18,846 (67\%) \\
\hline
\hline

\multirow{2}{*}{$S^D$} & $P = 0$ & 865/7,285 (12\%) & 5,811/7,285 (80\%) & 4,722/18,846 (25\%) & 10,603/18,846 (56\%) \\
\cline{2-6}
 & $P = 1$ & 0/7,285 (0\%) & 6,362/7,285 (87\%) & 0/18,846 (0\%) & 15,894/18,846 (84\%) \\

\hline
\hline
\Xhline{1.2pt}
\end{tabular}
\end{table}

To alleviate the aforementioned problem, we propose \baby by defining a novel supervised Dirichlet prior $\alpha'$, enabling to enrich discriminative supervision information. The novel prior $\alpha'$ consists of two components detailed below. 

\textbf{First}, we refine the prior $\alpha$ of LapSWTM to better describe the label membership degree. To achieve this, we consider not only the seed word occurrence information, but also the pseudo-nearest neighboring categories. Specifically, for each category, we compute a representative vector, referred to as prototype vector, by self-exploring the supervision information by word co-occurrences with seed words. For each document \emph{d}, we can compute its distances of all \emph{K} prototype vectors, hence considering that the document is associated with its \emph{P} pseudo-nearest neighboring categories. Built on this mechanism, formally named \textbf{prototype scheme}, for each label \emph{k} in document \emph{d} we jointly describe its membership degree $M_{dk}$ by the following formula:
\begin{equation}
\label{Eq4}
M_{dk} = (1-\tau) \frac {DF_{dk}}{\sum \nolimits _{i=1}^K DF_{di}} + \tau \frac{\mathcal{I}(k \in \Omega_d)}{P}, \quad \:\:  d=1,\cdots,D, \:\: k = 1,\cdots,K,
\end{equation}
where $\mathcal{I} (\cdot)$ is the indicator function; $\Omega _d$ is the pseudo-nearest neighboring category set of document \emph{d}; $\tau \in [0,1]$ is a tuning parameter used to balance the importance between the two components of Eq.\ref{Eq4}, and specifically note that $\sum \nolimits _{k=1}^K M_{dk}=1$.

Formally, we refer to label \emph{k} as \textbf{marked label} of document \emph{d} if $M_{dk} \not= 0$. The examples in Table \ref{Example} show us that the prototype scheme can effectively enhance the quality of the label membership degree prior. That is, more relevant labels can be covered by the marked labels with pseudo-nearest neighboring categories, naturally achieving supervision information enrichment.

\vspace{2pt}
\textbf{Second}, we also consider the label frequency prior at the corpus level, a commonly used discriminative knowledge in classification. Since the label frequency information is unknown, we must resort to an approximation. We assume that the label frequency is corresponding to its seed word occurrences. For each label \emph{k}, its label frequency prior $F_k$ can be computed by the following formula:
\begin{equation}
\label{Eq5}
F_k = \frac{\sum \nolimits _{v \in \mathbb{S}_k}TF_v}{\sum \nolimits _{i=1}^K \sum \nolimits _{v \in \mathbb{S}_i}TF_v }, \quad \quad \quad k=1,\cdots,K,
\end{equation}
where $\mathbb{S}_k$ denotes the seed word set of label \emph{k};  $TF_v$ is the number of times that word \emph{v} has occurred in the corpus, and additionally note that $\sum \nolimits _{k=1}^K F_k = 1$.

\vspace{2pt}
\noindent\textbf{Full formulation.} By combining Eqs.\ref{Eq4} and \ref{Eq5}, we suggest the supervised Dirichlet prior $\alpha'$ that simultaneously considers enriched label membership degree and label frequency priors. For any document \emph{d}, the \emph{k}th component of $\alpha'_d$ is defined to be:
\begin{equation}
\label{Eq6}
\alpha'_{dk} = \eta\left(\left(1-\rho\right) M_{dk} + \rho F_k\right) + \alpha_0, \quad \quad d=1,\cdots,D, \:\: k=1,\cdots,K,
\end{equation}
where $\rho \in [0,1]$ is a tuning parameter used to balance the importance between $M_{dk}$ and $F_k$; $\eta$ is a concentration parameter; and $\alpha_0$ is a smoothing parameter used to avoid zero.

\begin{table}
\footnotesize
\caption{Examples of supervised Dirichlet priors of \emph{Reuters} documents (with seed word set $S^L$) computed by LapSWTM of Eq.\ref{Eq1} (\ie $\alpha$) and \baby of Eq.\ref{Eq6} (\ie $\alpha'$). The same parameters of Eqs.\ref{Eq1} and \ref{Eq6} used here are fixed as $\eta=10$ and $\alpha_0=0.01$, and the specific parameters of $\alpha'$ are fixed as $P=1$ (\ie the number of pseudo-nearest neighboring categories in Eq.\ref{Eq4}) and $\tau=0.1$ (\ie the tuning parameter in Eq.\ref{Eq4}).}   
    \centering
    \begin{tabular}{p{100pt}<{\centering}| p{40pt}<{\centering}| p{200pt}<{\centering}}
        \Xhline{1.5pt}

        Seed word set   & \multicolumn{2}{c}{Label (Seed word)} \\
        \Xhline{1pt}
        \multirow{3}{*}{$S^L$ of \emph{Reuters}} & \multicolumn{2}{c}{\multirow{3}{210pt}{ \:\: 1 (acquisition)\:\: 2 (coffee)\:\: 3 (crude)\:\: 4 (earnings)\:\: 5 (gold) \quad\quad 6 (interest)\:\: 7 (foreign,\, exchange)\:\: 8 (ship)\:\: 9 (sugar)\:\: 10 (trade)}} \\
        & \multicolumn{2}{p{248pt}<{\centering}}{} \\
        & \multicolumn{2}{p{248pt}<{\centering}}{} \\
        \Xhline{1pt}
        (Label) Document  & Prior   & Prior value \\
        \Xhline{1pt}
        \multirow{9}{100pt}{\textcolor{blue}{(Label: 1)} \quad terminal sale mln did dlrs company gulf applied technologies sells units subsidiaries engaged pipeline operations 12 subject post closing adjustments}  
        & \multirow{3}{*}{$\alpha$} & \multirow{3}{*}{$[\,\textcolor{blue}{0.01},0.01,0.01,0.01,0.01,0.01,0.01,0.01,0.01,0.01\,]$}\\
        &  &  \\
        &  &  \\
        & \multirow{3}{*}{$\alpha'$ ($\rho$ = 0)} & \multirow{3}{*}{$[\,\textcolor{blue}{10.01},0.01,0.01,0.01,0.01,0.01,0.01,0.01,0.01,0.01\,]$} \\
        &  &  \\
        &  &  \\
        & \multirow{3}{*}{$\alpha'$ ($\rho$ = 0.9)} & \multirow{3}{*}{$[\,\textcolor{blue}{3.54},0.15,0.41,4.60,0.12,0.25,0.31,0.19,0.15,0.38\,]$} \\
        &  &  \\
        &  &  \\
        \hline
        \multirow{9}{100pt}{\textcolor{blue}{(Label: 1)} \quad systems 000 shares common bought maker 677 272 following 969 643 holders 74 boards revenues pretax 232 204 pct outstanding circuit profits 1986 stock buys industries mln \textcolor{red}{exchange} dlrs 1985 board}  
        & \multirow{3}{*}{$\alpha$} & \multirow{3}{*}{$[\,\textcolor{blue}{0.01},0.01,0.01,0.01,0.01,0.01,\textcolor{red}{10.01},0.01,0.01,0.01\,]$} \\
        &  &  \\
        &  &  \\
        & \multirow{3}{*}{$\alpha'$ ($\rho$ = 0)} & \multirow{3}{*}{$[\,\textcolor{blue}{1.01},0.01,0.01,0.01,0.01,0.01,\textcolor{red}{9.01},0.01,0.01,0.01\,]$} \\
        &  &  \\
        &  &  \\
        & \multirow{3}{*}{$\alpha'$ ($\rho$ = 0.9)} & \multirow{3}{*}{$[\,\textcolor{blue}{2.64},0.15,0.41,0.460,0.12,0.25,\textcolor{red}{0.121},0.19,0.15,0.38\,]$} \\
        &  &  \\
        &  &  \\

        \Xhline{1.5pt}
    \end{tabular}
    \label{ExampleOfPrior}
\end{table}

\vspace{2pt}
\noindent\textbf{Visualization of supervised Dirichlet prior}: To further describe the novel supervised Dirichlet prior (\ie Eq.\ref{Eq6}), we provide some examples in Table \ref{ExampleOfPrior}. Accordingly, we illustrate the seed word set $S^L$ of \emph{Reuters} and two example documents associated with label 1. As shown in Table \ref{ExampleOfPrior}, we have the following observations on the supervised Dirichlet priors computed by different methods. (1) Note that the first example document contains no seed words, the prior $\alpha$ of LapSWTM computed by Eq.\ref{Eq1} will be a uniform prior without any supervision for this document. In contrast, our novel prior $\alpha' (\rho=0)$ effectively incorporates accurate supervision on the ground-truth label by leveraging the prototype scheme. Besides, the prior $\alpha' (\rho=0.9)$ further considers the label frequency information, giving more discriminative supervised signals for the document. (2) We can see that the second example document only contains an irrelevant seed word of label 7 but no seed words of the ground-truth label, \ie label 1, resulting in a noisy supervised prior $\alpha$. For this situation, our novel prior $\alpha' (\rho=0)$ with the prototype scheme can also revise the prior $\alpha$ to some extent. The prior $\alpha' (\rho=0.9)$ is again a more discriminative supervised prior by further considering the label frequency information. 

\subsubsection{Computation of Prototype Vectors}

The prototype scheme requires discriminative prototype vectors to guarantee the effective discovery of unobservable label membership degree prior. To this end, we utilize word co-occurrences with seed words to self-explore more supervision information. Inspired by \cite{CFC2009}, we compute the \emph{v}th component of the prototype vector of category \emph{k} by:
\begin{equation}
\label{Eq7}
c_{kv} = b ^{\frac{SF_{kv}}{S_k}} \times \ln \left(\frac{K}{CF_{v}}\right), \quad \quad k = 1,\cdots,K, \:\: v= 1, \cdots,V,
\end{equation}
where $SF_{kv}$ is the number of co-occurrences\footnote{Here, word co-occurrence means two words co-occur in the same document.} between word \emph{v} and seed words of category \emph{k}; $S_k$ is the total number of times that seed words of category \emph{k} have occurred; $CF_v$ is the number of categories containing word \emph{v} measured by word co-occurrences with seed words; \emph{b} is a constant fixed by $e-1$ in this work.

The first term of Eq.\ref{Eq7} represents the importance of word \emph{v} in category \emph{k}, and the second term describes the discriminative power of word \emph{v} among different categories. That is, this prototype vector considers both the inner-category word index and inter-category word index, which can effectively embed the relevant labels of documents into the top-\emph{P} neighboring label sets.

Additionally, in this work, all documents are represented by term frequency vectors, and the cosine similarity measurement is employed to find the top-\emph{P} neighboring categories.

\subsubsection{Term Weighting}

In the context of dataless text classification, every word token is important since it lacks label information at the document level. Less discriminative words, \eg domain-specific stopwords, must make the model less effective \cite{IPM2018}. To alleviate this problem, we employ a term weighting scheme to filter out such words. Here, we compute the word weight using the Log weight \cite{WLDA2010}, supposing that high frequency words contribute little to the semantics of documents, such as the stopwords ``an'' and ``the''. It punishes such words following the axiom of information theory that the information content of an event a is equivalent to $-\log p(a)$. Therefore, the Log weight can be computed by:
\begin{equation}
\label{Eq8}
\pi(v) = - \log p(v), \quad \quad \quad v=1,\cdots,V,
\end{equation}
where $p(v)$ is estimated by the occurrences of word \emph{v} in the corpus.

\subsection{Inference for \baby}
\label{3.3}

The models of LapSWTM and \baby share the same model structure, but in \baby we utilize a novel supervised Dirichlet prior $\alpha'$ (\ie Eq.\ref{Eq6}) and weigh words by the Log weight (\ie Eq.\ref{Eq8}). Therefore, we can infer \baby by applying the same inference process of LapSWTM, after computing $\alpha'$ and considering the word weights as the soft occurrence numbers of words.\footnote{That means once a word token $w_{dn}$ is observed, this word is considered as occurring $\pi(w_{dn})$ times.} For clarity, we briefly present the inference process details by referring to \cite{CIKM2018}.

We first compute the supervised Dirichlet prior $\alpha'$ and the word weights $\pi$, and then apply generalized expectation maximization to maximize the following objective with respect to $\{\theta,\widehat \theta, \phi, \widehat \phi\}$:
\begin{equation}
\label{Eq9}
\mathcal{L}'(\theta,\widehat \theta,\phi,\widehat \phi) = \log p\left(\mathbb{W},\theta,\widehat \theta,\phi,\widehat \phi|\alpha',\widehat \alpha,\beta,\widehat \beta,\gamma,\pi\right) - \lambda \mathcal{R}(\theta)
\end{equation}
Naturally, this can be achieved by iteratively applying the following \textbf{E-step} and \textbf{M-step} until convergence. The derivations of key update equations are detailed in the Appendix. 

\begin{algorithm}[t]
  \caption{Newton-Raphson loop for $\theta$}
  \begin{algorithmic}[1]
    \State ``Initialize'' $\theta$ using Eq.\ref{Eq15}
    \State Copy $\theta$ to a temporary variable $\theta'$
    \State \textbf{While} $\mathcal{L}(\theta) \le \mathcal{L}(\theta')$ \textbf{Do}
    \State $\quad$ Copy $\theta'$ to $\theta$
    \State $\quad$ Update $\theta'$ using Eq.\ref{Eq16}
    \State \textbf{End While}
   \end{algorithmic}
\end{algorithm}

\begin{algorithm}[t]
  \caption{Inference for \baby}
  \begin{algorithmic}[1]
    \State Initialize parameters and hidden variables randomly
    \State Build the document graph for manifold regularizer
    \State Compute $\alpha'$ using Eq.\ref{Eq6}
    \State Compute $\pi$ using Eq.\ref{Eq8}
    \State \textbf{For} $t = 1,2,\ldots,$ MaxIter
    \State $\quad$ \textbf{E-step}: Update $\{N_{dnk}\}_{k=1}^K, \{\widehat N_{dng}\}_{g=1}^G$ for each word token by Eqs.\ref{Eq10} and \ref{Eq11}
    \State $\quad$ \textbf{M-step}:
    \State $\quad$ $\quad$ Update $\{\widehat \theta, \phi, \widehat \phi\}$ by Eqs.12, 13 and 14
    \State $\quad$ $\quad$ Update $\theta$ using \textbf{\emph{Algorithm 1}}
    \State \textbf{End for}
   \end{algorithmic}
\end{algorithm}

\vspace{0.5ex}
{\textbf{[E-step]}}: 
Given the current $\{\theta,\widehat \theta, \phi, \widehat \phi\}$, for each word token $w_{dn}$ we estimate the posterior of topic assignment by applying the Bayes rule:
\begin{equation}
\label{Eq10}
p\left(z_{dn}=k\right) = \theta_d^T\gamma_{w_{dn}} \frac{\theta_{dk}\phi_{kw_{dn}}}{\sum \nolimits_{i=1}^K \theta_{di}\phi_{iw_{dn}}} \buildrel \Delta \over = N_{dnk}, \quad \quad k=1,\cdots, K,
\end{equation}
\begin{equation}
\label{Eq11}
p\left(\widehat z_{dn}=g\right) = \left(1-\theta_d^T\gamma_{w_{dn}}\right) \frac{\widehat \theta_{dg}\widehat \phi_{gw_{dn}}}{\sum \nolimits_{i=1}^G \widehat \theta_{di}\phi_{iw_{dn}}} \buildrel \Delta \over = \widehat N_{dng}, \quad \quad g=1,\cdots,G
\end{equation}
For convenience, we denote by $N_{dnk}$ and $\widehat N_{dng}$ the posteriors of category topic \emph{k} and background topic \emph{g} for word token $w_{dn}$, respectively.

\vspace{2ex}
{\textbf{[M-step]}}: Given the current $\{N_{dnk}\}_{k=1}^K, \{\widehat N_{dng}\}_{g=1}^G$ of all word tokens, we update the latent variables of interest, \ie $\{\theta,\widehat \theta, \phi, \widehat \phi\}$.

For $\widehat \theta$, $\phi$ and $\widehat \phi$ without manifold regularizer, they can be directly updated by the following equations with word weights:
\begin{eqnarray}
\label{Eq121314}
\widehat \theta_{dg} = \frac{\sum \nolimits_{n=1}^{N_d}\pi(w_{dn})\widehat N_{dng}+\widehat \alpha}{\sum \nolimits _{i=1}^G \sum \nolimits_{n=1}^{N_d}\pi(w_{dn})\widehat N_{dni}+ G\widehat \alpha} \:,\\
\phi_{kv} = \frac{\sum \nolimits_{d=1}^{D} \sum \nolimits_{v \in d} \pi(v) N_{dnk}+\beta}{\sum \nolimits_{d=1}^{D} \sum \nolimits_{n=1}^{N_d} \pi(w_{dn})N_{dnk} + V \beta} \:,\\
\widehat \phi_{gv} = \frac{\sum \nolimits_{d=1}^{D} \sum \nolimits_{v \in d} \pi(v) \widehat N_{dng}+ \widehat \beta}{\sum \nolimits_{d=1}^{D} \sum \nolimits_{n=1}^{N_d} \pi(w_{dn}) \widehat N_{dng} + V \widehat \beta}\:,
\end{eqnarray}
where $N_d$ denotes the number of word tokens in document $d$.

For $\theta$, we update it using an inner Newton-Raphson loop on $\mathcal{R}(\theta)$ until the objective of Eq.\ref{Eq9} decreases. In this inner loop, we ``initialize'' $\theta$ by the following equation:
\begin{equation}
\label{Eq15}
\theta_{dk} = \frac{\sum \nolimits_{n=1}^{N_d}\pi(w_{dn})N_{dnk}+ \alpha'_{dk}}{\sum \nolimits _{i=1}^K \sum \nolimits_{n=1}^{N_d}\pi(w_{dn})N_{dni}+ \sum \nolimits_{i=1}^{K} \alpha'_{di}}
\end{equation}
and then iteratively update it as follows:
\begin{equation}
\label{Eq16}
\theta_{dk} \leftarrow (1-\kappa)\theta_{dk} + \kappa \frac{\sum \nolimits_{i=1}^D\theta_{ik} W_{di}}{\sum \nolimits_{i=1}^DW_{di}},
\end{equation}
where $\kappa \in [0,1]$ is the step size. For clarity, we outline this inner Newton-Raphson loop in \textbf{\emph{Algorithm 1}}, and the full inference process of \baby in \textbf{\emph{Algorithm 2}}.

\section{Experiment}
\label{4}


\subsection{Experimental Setting}
\label{4.1}

\vspace{2ex}
\noindent{\textbf{Datasets.}}
We employed two datasets used in \cite{STM2016,CIKM2018}, including \emph{Reuters}\footnote{http://kdd.ics.uci.edu/database/reuters21578/reuters21578.html} and \emph{Newsgroup}.\footnote{http://qwone.com/$\sim$jason/20Newsgroups/} The original \emph{Reuters} dataset contains 135 categories, which are extremely \textit{imbalanced}. We left the largest 10 categories, consisting of 7,285 documents in total, where 5,228 documents for training and 2,057 documents for testing. We selected the \emph{bydate} version of \emph{Newsgroup} that contains totally 18,846 documents in 20 categories. We used the standard train/test split, where the training and test sets contain 11,314 and 7,532 documents, respectively. \emph{Newsgroup} is a \textit{balanced} dataset. Preprocessing steps include removals of the standard stopwords, the words shorter than 2 characters and appearing less than 5 documents.

Following \cite{AAAI2015,STM2016,CIKM2018}, we used two publicly available seed word sets, denoted by $S^L$ and $S^D$. The seed words of $S^L$ come from the category descriptions by further removing unrelated words. The seed words of $S^D$ are extracted by unsupervised methods with further manual selection. $S^L$ contains only a few seed words, while the volume of $S^D$ is relatively larger. Besides, for $S^D$ of \emph{Newsgroup} some seed words are shared by different categories. Here, we purify it by randomly retaining each overlapping seed word to one of its categories. The statistics of datasets and seed word sets are summarized in Table \ref{dataset}.

\vspace{2ex}
\noindent{\textbf{Baseline methods.}}
We employed four existing dataless methods for comparison. Besides them, we also compare \baby against two supervised methods trained over labeled documents.

\begin{itemize}
    \item \textbf{PL-DNB} \cite{coling2018} is a discriminative dataless method, built on naive Bayes. In the experiments, we use its code supplied by its authors. We run PL-DNB with different parameter configurations and leave the best scores for comparison.
    
    \item \textbf{STM} \cite{STM2016} is a topic modeling based dataless method using seed words of categories. We train this model using the public code\footnote{https://github.com/ly233/Seed-Guided-Topic-Model} implemented by its authors, and set its parameters following the suggestions in its original paper.
    
    \item \textbf{LapSWTM} \cite{CIKM2018} is the ancestor method of our \baby, where the main difference between them is the construction ways of the supervised Dirichlet prior for category topics. In the experiments, we use its code supplied by its authors and set its parameters following the suggestions in its original paper.
    
    \item \textbf{WeSTClass} \cite{Mengyu2018} is a deep learning dataless method. We train this model using the public code\footnote{https://github.com/yumeng5/WeSTClass} implemented by its authors, and set its parameters following the suggestions in its original paper.
    
    \item \textbf{Support vector machines} (\textbf{SVMs}) is a traditional discriminative supervised method. Here, we represent documents as TF-IDF features and train a SVMs classifier using the \emph{sklearn} tool\footnote{http://scikit-learn.org/stable/} with default parameter settings.
    
    \item \textbf{Maximum entropy discrimination latent Dirichlet allocation} (\textbf{MedLDA}) \cite{MedLDA2012} is a traditional supervised topic model for classification. We train this model using the public code\footnote{http://ml.cs.tsinghua.edu.cn/$\sim$jun/gibbs-medlda.shtml} implemented by its authors. Specially, the option of inner cross-validation optimization is applied.
\end{itemize}

\begin{table}[t]
\centering
\small
\caption{Statistics of datasets and seed word sets. \emph{D}: number of documents; \emph{V}: vocabulary size; \emph{K}: number of categories; $|S^L|$ and $|S^D|$: average numbers of seed words in $S^L$ and $S^D$, respectively.}
\label{dataset}
\renewcommand\arraystretch{1.35}
\begin{tabular}{p{2cm}<{\centering}|p{1.2cm}<{\centering}p{1.2cm}<{\centering}p{0.9cm}<{\centering}p{0.9cm}<{\centering}p{0.9cm}<{\centering}}
\Xhline{1.5pt}
Dataset     &\emph{D}  &\emph{V}  &\emph{K} & $|S^L|$ & $|S^D|$ \\
\Xhline{1.2pt}
\emph{Reuters}          &7,285	  &7,418	     & 10	  & 1.1    &6   \\ \hline
\emph{Newsgroup}         &18,846   &52,761    &20    & 1.5  &4.75 \\
\Xhline{1.5pt}
\end{tabular}
\end{table}

For our \textbf{\baby}, its parameter setting follows that of LapSWTM. Additionally, the specific parameters used to compute $\alpha'$ are set as follows: $\eta=10$, $\alpha_0=0.01$, $P=1$, $\rho=0.9$ and $\tau=0.1$

Besides, we emphasize that following \cite{STM2016,CIKM2018}, for all dataless methods we train them on both training and test documents as a single collection, and report the classification performance evaluated on the test documents.

\vspace{2ex}
\noindent{\textbf{Evaluation metrics.}}
We measure the classification performance by employing Micro-F1 and Macro-F1, where Micro-F1 is the F1 score over the dataset and Macro-F1 is the average of the F1 scores within categories.

\subsection{Classification Performance Comparison}
\label{4.2}

For all methods, we perform 10 independent runs and report the average results. The Micro-F1 and Macro-F1 scores are shown in Tables \ref{TableOfMicroF1} and \ref{TableOfMacroF1}, respectively. We made the following observations.

\vspace{1ex}
\noindent{\textbf{Comparing with dataless methods}}:
Overall, we can observe that \baby significantly outperforms existing dataless methods, and \baby+$S^D$ achieves the highest scores in all the cases. For PL-DNB and STM, the performance gains of \baby are relatively significant, where the Micro-F1 and Macro-F1 scores of \baby are about $0.05\sim0.1$ and $0.06\sim0.09$ higher than those of the two baseline methods, respectively. More importantly, it can be seen that \baby outperforms LapSWTM and WeSTClass, where the performance improvements are about $0.01\sim0.06$. Note that he main difference between LapSWTM and \baby is the formulation of the supervised Dirichlet prior of category topics. Therefore the performance gap over LapSWTM indicates the effectiveness of the proposed prior of \baby, indirectly implying that it can effectively enrich and modify the supervised signals from seed words..

We now discuss the results of different seed word sets. First, we can observe that all methods with $S^D$ perform better than the ones with $S^L$ on both datasets. This is consistent to our expectation, since $S^D$ contains more seed words than $S^L$, giving more supervision information. Besides, we can observe that the performance gap of \baby between $S^D$ and $S^L$ is smaller than those of baseline methods. For example, the Macro-F1 score of \baby+$S^D$ is about 0.04 higher than that of \baby+$S^L$ on \emph{Newsgroup}, but the gaps of PL-DNB, STM, LapSWTM, and WeSTClass are about 0.06, 0.07, 0.09, and 0.06, respectively. The results indicate that \baby is less sensitive to the number of seed words, while it can achieve competitive performance with fewer seed words. This makes \baby more practical, since collecting seed words may be still a difficult task in domain-specific real applications. Comparing between LapSWTM and \baby, we can observe that \baby performs about $0.01\sim0.02$ higher scores on $S^D$, while about $0.03\sim0.06$ higher scores on $S^L$. First, the performance improvement on $S^D$ is less obvious. The possible reason is that given a larger seed word set one can compute relatively accurate category prior using only seed word occurrences as doings in LapSWTM. Second, \baby significantly improves both Micro-F1 and Macro-F1 scores on $S^L$, hence implying that the proposed category prior works well with even few seed words. This again indicates that \baby is more practical in real applications.

\begin{table*}[t]
\renewcommand\arraystretch{2.0}
\caption{Experimental results of Micro-F1. The notation ``$\ddag$'' means that the gain of \baby+$S^D$ is statistically significant at 0.01 level (paired sample t-test).}
\label{TableOfMicroF1}
\centering
\footnotesize
\begin{tabular}{m{36pt}<{\centering}|m{10pt}<{\centering}|m{30pt}<{\centering}|m{30pt}<{\centering}|m{30pt}<{\centering}|m{38pt}<{\centering}|m{34pt}<{\centering}|m{24pt}<{\centering}|m{32pt}<{\centering}}
\Xhline{1.2pt}
\multicolumn{2}{c|}{Dataset}  & \textbf{\baby}  & \textbf{PL-DNB}   &  \textbf{STM}  &  \textbf{LapSWTM}  & \textbf{WeSTClass}   &  \textbf{SVMs} &  \textbf{MedLDA} \\ \cline{2-9}
\Xhline{1.2pt}
\multirow{2}{*}{\emph{Reuters}} 
& $S^L$   &0.938   &0.849$^\ddag$   &0.859$^\ddag$   & 0.877$^\ddag$  & 0.879$^\ddag$  & \multirow{2}{*}{0.973}   & \multirow{2}{*}{0.934} \\ \cline{2-7}
& $S^D$   &\textbf{0.952}   &0.891$^\ddag$  &0.901$^\ddag$  &0.931$^\ddag$  & 0.913$^\ddag$  &  & \\ \hline

\multirow{2}{*}{\emph{Newsgroup}}  
& $S^L$ &0.805   &0.715$^\ddag$   &0.699$^\ddag$   &0.748$^\ddag$  & 0.758$^\ddag$  & \multirow{2}{*}{0.823}   & \multirow{2}{*}{0.811} \\ \cline{2-7}
& $S^D$ &\textbf{0.822}   &0.765$^\ddag$   &0.771$^\ddag$   &0.810$^\ddag$  & 0.802$^\ddag$   & & \\
\Xhline{1.2pt}
\end{tabular}
\end{table*}

\begin{table*}[t]
\renewcommand\arraystretch{2.0}
\caption{Experimental results of Macro-F1. The notation ``$\ddag$'' means that the gain of \baby+$S^D$ is statistically significant at 0.01 level (paired sample t-test).}
\label{TableOfMacroF1}
\centering
\footnotesize
\begin{tabular}{m{36pt}<{\centering}|m{10pt}<{\centering}|m{30pt}<{\centering}|m{30pt}<{\centering}|m{30pt}<{\centering}|m{38pt}<{\centering}|m{34pt}<{\centering}|m{24pt}<{\centering}|m{32pt}<{\centering}}
\Xhline{1.2pt}
\multicolumn{2}{c|}{Dataset}  & \textbf{\baby}  & \textbf{PL-DNB}   &  \textbf{STM}  &  \textbf{LapSWTM}  & \textbf{WeSTClass}   &  \textbf{SVMs} &  \textbf{MedLDA} \\ \cline{2-9}
\Xhline{1.2pt}
\multirow{2}{*}{\emph{Reuters}} 
& $S^L$   &0.830   &0.758$^\ddag$   &0.753$^\ddag$   &0.799$^\ddag$    & 0.781$^\ddag$  & \multirow{2}{*}{0.939}   & \multirow{2}{*}{0.916} \\ \cline{2-7}
& $S^D$   &\textbf{0.900}   &0.837$^\ddag$   &0.835$^\ddag$  &0.877$^\ddag$  & 0.853$^\ddag$ &  & \\ \hline

\multirow{2}{*}{\emph{Newsgroup}}  
& $S^L$ & 0.759   & 0.677$^\ddag$   & 0.669$^\ddag$   & 0.693$^\ddag$    & 0.711$^\ddag$  & \multirow{2}{*}{0.816}   & \multirow{2}{*}{0.802} \\ \cline{2-7}
& $S^D$ & \textbf{0.798}   & 0.731$^\ddag$    & 0.737$^\ddag$   & 0.783$^\ddag$  & 0.779$^\ddag$   & & \\

\Xhline{1.2pt}
\end{tabular}
\end{table*}

\begin{table}[t]
\centering
\caption{Summary of the Friedman statistics $F_F$ (l=5, n=40) and the critical value of each evaluation metric.}
\renewcommand\arraystretch{1.2}
\begin{tabular}{p{100pt}<{\centering} p{40pt}<{\centering} p{120pt}<{\centering}}
\Xhline{1.5pt}

Evaluation metric & \emph{$F_F$}  & Critical Value ($\alpha$ = 0.05) \\
\Xhline{0.5pt}

Micro-F1  & 247.5014     & \multicolumn{1}{c}{\multirow{2}{*}{2.4296}} \\
Macro-F1  & 409.2759   \\

\Xhline{1.5pt}

\end{tabular}
\label{Table_Friedman}
\end{table}

\begin{figure*}[t]
\includegraphics[width=0.95\textwidth]{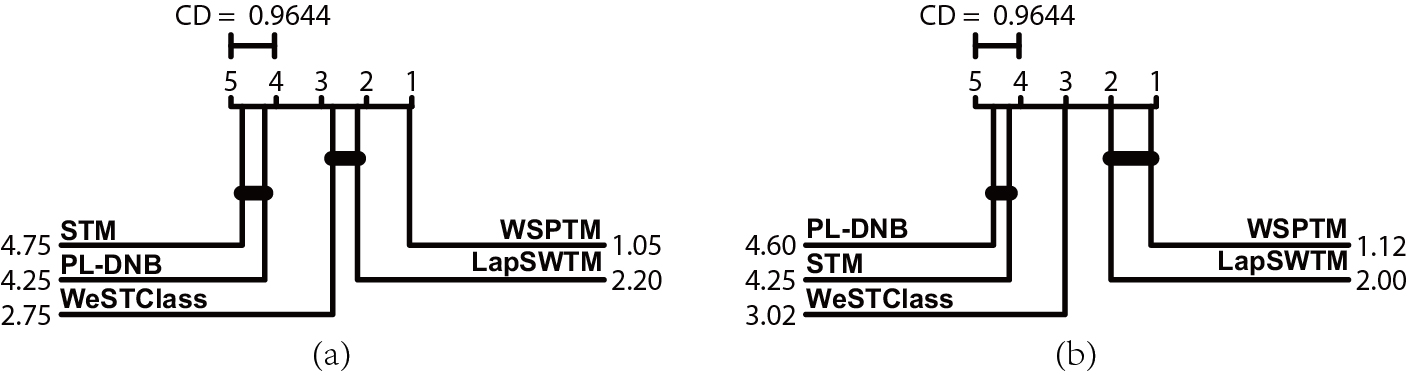}
\centering
\caption{Comparisons of the average ranks of totally $40$ evaluation results using the Nemenyi test on each evaluation metric: (a) \textbf{Micro-F1}, (b) \textbf{Macro-F1}. The comparing methods that are not significantly different at 0.05 level (\ie \emph{p} = 0.05) are connected.}
\label{Fig1}
\end{figure*}

\vspace{1ex}
\noindent{\textbf{Comparing with supervised methods}}:
We now compare \baby against SVMs and MedLDA. Overall, we can see that \baby+$S^D$ performs a bit worse than SVMs, and achieves very competitive performance with MedLDA. The Micro-F1 and Macro-F1 scores of \baby+$S^D$ are only about $0.001\sim0.04$ lower than those of SVMs. Surprisingly, \baby+$S^D$ gets higher Micro-F1 scores than MedLDA across both datasets. In contrast to the supervised methods, \baby trains a classifier using only very small numbers of seed words, instead of the expensive labeled documents, so as to save many human efforts. In this sense, we kindly argue that \baby can be regarded as an effective supplement to the traditional supervised methods.

Additionally, our early experiments have indicated that \baby+$S^D$ outperformed several other supervised methods, such as \emph{K}NN and supervised LDA \cite{sLDA2007}, in the most cases. The results further indicate the effectiveness of \baby.

\vspace{1ex}
\noindent \textbf{Statistical analysis:} 
We employ the Friedman test \cite{Demsar2006} to statistically analyze the relative performance among all comparing dataless methods. For each evaluation metric (\ie Micro-F1 and Macro-F1), we report the Friedman test statistics $F_F$ (\#comparing methods: $l$=5 and \#evaluation results: $n$=40, \ie 2 datasets with 2 seed word sets given 10 independent runs) and the critical value in Table \ref{Table_Friedman}. Obviously, the null hypothesis that all comparing methods perform equivalently is rejected at the significance level $\alpha \, = \, 0.05$.

To further analyze the relative performance among the comparing dataless methods, we adopt the Nemenyi test \cite{Demsar2006}, where our \baby is deemed as the control method. For each evaluation metric, we independently rank all comparing methods over totally $40$ evaluation results, and compare the average ranks of pairwise comparing methods. Specifically, the performance between two methods will be considered significantly different if the corresponding average ranks differ by at least the critical difference (CD), formulated as follows \cite{Demsar2006}:
\begin{equation}
    CD = q_a\sqrt{\frac{l(l+1)}{6n}},
\end{equation}
where here we have $l=5$, $n=40$ and $q_a = 2.7278$ at significance level $\alpha = 0.05$, leading to $\rm{CD} = 0.9644$. The CD diagrams are shown in Fig.\ref{Fig1}. We can observe that \baby ranks the first statistically on both evaluation metrics. Besides, LapSWTM and WeSTClass rank the second and third. and they performs significantly better than other PL-DNB and STM.

\subsection{Evaluation of Held-out Likelihood}

We compare the held-out likelihood performance between LapSWTM and \baby. For both models, we estimate the optimal category topics $\phi^*$ and background topics $\widehat \phi^*$, and then compute the perplexity scores over the test documents $\mathbb{\widehat W} = \{\widehat w_d\}_{d=1}^{\widehat D}$ as follows:  
\begin{eqnarray}
    {\rm{Perplexity}}(\mathbb{\widehat W}) = \exp \left(-\:\: \frac{\sum_{d=1}^{\widehat D}\log\left(p\left(\widehat w_d| \alpha,\widehat \alpha,\phi^*,\widehat \phi^*,\gamma,\pi\right)\right)}{\sum_{d=1}^{\widehat D}N_d}\right), \nonumber
\end{eqnarray}
During testing, the manifold regularization is not applied for both models. The results are shown in Table \ref{Perplexity}. We can observe that \baby consistently performs better than LapSWTM, indicating that \baby can better model the text document collections.

\begin{table}[t]
  \centering
  \caption{Experimental results of perplexity.}
  \renewcommand\arraystretch{1.35}
  \small
  \begin{tabular}
  {p{60pt}<{\centering}|p{45pt}<{\centering}p{45pt}<{\centering}}
    \Xhline{1.5pt}
    Dataset &  \baby & LapSWTM   \\
    \Xhline{1pt}

	\multirow{1}{*}{Reuters} 
	&\textbf{{1572.5}}	&{1628.3}\\
	\hline
	
	\multirow{1}{*}{Newsgroup} 
	&\textbf{3024.1}	&{3216.3}	\\
	\hline
	
	\hline
    \Xhline{1.2pt}
  \end{tabular}
  \label{Perplexity}
\end{table}

\subsection{Evaluation on Parameters}
\label{4.3}

We now empirically evaluate the crucial parameters of \baby, including $\rho$, $\tau$ and \emph{P}.

\subsubsection{Evaluation of Tuning Parameters $\rho$ and $\tau$}
\label{4.3.1}

\begin{figure*}[t]
\includegraphics[width=0.95\textwidth]{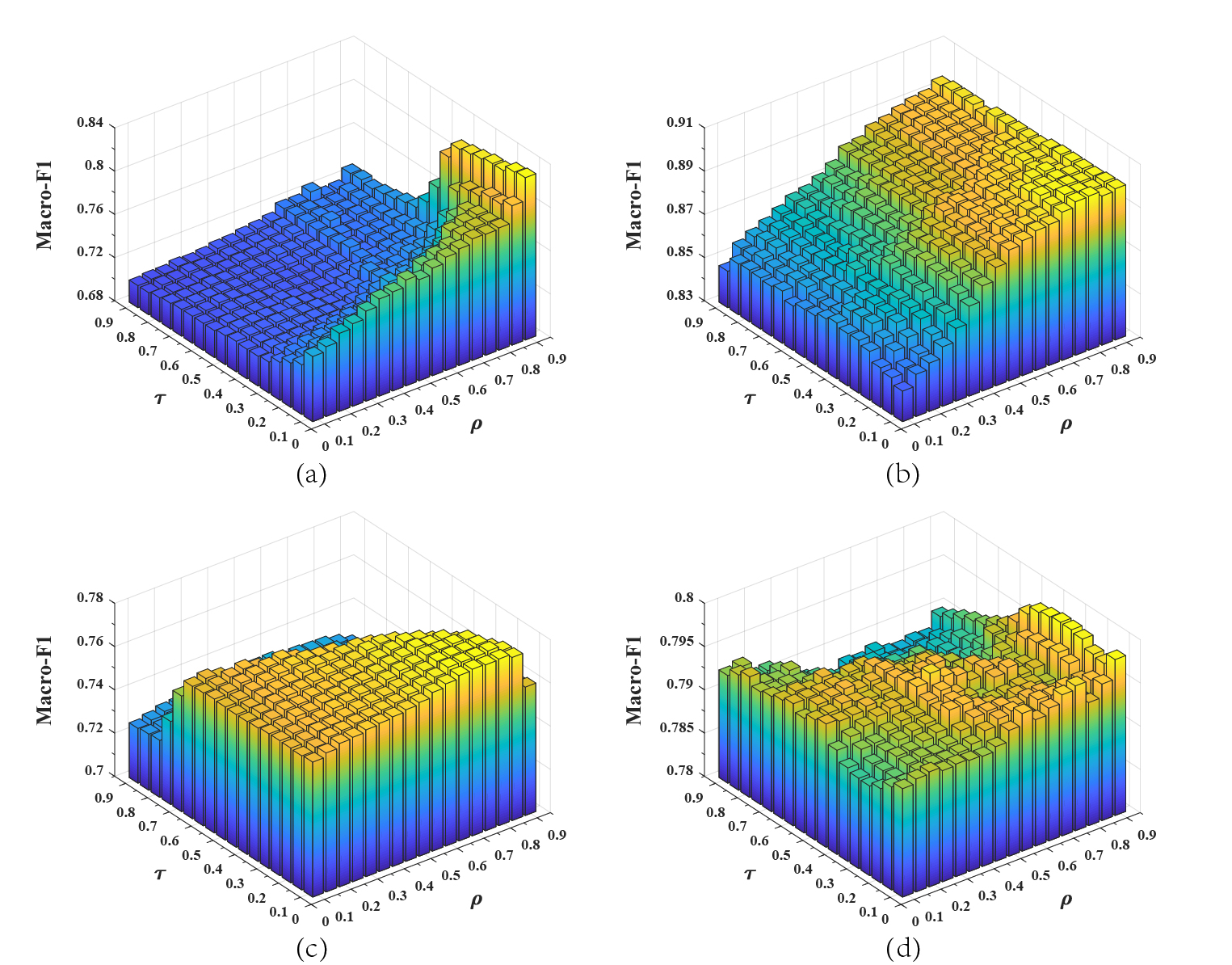}
\centering
\caption{The Macro-F1 scores of different $\rho$ and $\tau$ values on (a) \emph{Reuters} with $S^L$, (b) \emph{Reuters} with $S^D$, (c) \emph{Newsgroup} with $S^L$ and (d) \emph{Newsgroup} with $S^D$}
\label{Fig2}
\end{figure*}

This subsection shows the empirical evaluation results of the two tuning parameters over the set $\{0.1,0.15,\cdots,0.9\}$, and Figure \ref{Fig2} plots the experimental results of Macro-F1 scores by fixing all other parameters. We now discuss the two parameters one by one.

Reviewing Eq.\ref{Eq6}, the parameter $\rho$ is used to control the importance between the membership degree of labels and label frequencies in $\alpha'$, while the value of $\rho$ corresponds to the proportion of the label frequency prior. In terms of the imbalanced \emph{Reuters} dataset, we can observe that \baby consistently performs better as the value of $\rho$ increases. We argue that is because for the imbalanced dataset such as \emph{Reuters} the label frequency prior is a significantly discriminative knowledge for classification. In terms of the balanced \emph{Newsgroup} dataset, the performance carves of \baby are relatively stable. Especially when $S^D$ is used, the performance gaps between different $\rho$ values are almost less than 0.01. This result matches our expectations, where the balanced dataset such as \emph{Newsgroup} may be insensitive to the label frequency prior.

Reviewing Eq.\ref{Eq4}, the parameter $\tau$ is used to describe the importance between seed word occurrences and pseudo-nearest neighboring
categories in the label membership degree prior, while the value of $\tau$ describes the proportion of pseudo-nearest neighboring categories. First, we observe that \baby performs stable when $S^D$ is used. Second, the Macro-F1 scores of \baby are gradually decreasing as the value of $\tau$ increases when $S^L$ is used. For example, the Macro-F1 scores become much lower on \emph{Newsgroup} when $\rho > 0.6$. The possible reason is that given fewer seed words we can obtain less accurate pseudo-nearest neighboring categories, which are depended on self-exploring the available data with seed words. Larger values of $\tau$ may introduce additional noise, reducing the classification performance. Therefore, we prefer smaller values of $\tau$ for safer results.

\vspace{1ex}
\noindent \textbf{Ablative study:} To clearly show the impact of the two components of the proposed category prior, we specifically examine them by removing pseudo-nearest neighboring categories (PNNC) and label frequencies (LF). The Macro-F1 scores of different versions are shown in Table \ref{Ablation}. It is clearly seen that both components provide positive effects on improving Macro-F1. The label frequency has significant impact on \textit{Reuters}, due to its imbalance. 



\begin{table*}[t]
\renewcommand\arraystretch{1.35}
\caption{The Macro-f1 scores of ablative study.}
\label{Ablation}
\centering
\footnotesize
\begin{tabular}{m{48pt}<{\centering}|m{45pt}<{\centering}|m{45pt}<{\centering}|m{45pt}<{\centering}|m{45pt}<{\centering}}
\Xhline{1.2pt}
  & \multicolumn{2}{c|}{\textit{Reuters}}    & \multicolumn{2}{c}{\textit{Newsgroup}}   \\ 
\cline{2-5}
  & $S^L$   & $S^D$    & $S^L$   & $S^D$   \\
\Xhline{1.2pt}

\textbf{\baby}  & \textbf{0.830}    &\textbf{0.900}    &\textbf{0.759}     &\textbf{0.798}  \\

\hline

\textbf{-PNNC}              &0.778    &0.890    & 0.723     &0.790  \\ \hline

\textbf{-LF}            & 0.733   &0.839   &0.751  &0.796   \\ \hline


\Xhline{1.2pt}
\end{tabular}
\end{table*}

\subsubsection{Evaluation of Pseudo-nearest Label Number P}
\label{4.3.2}

\begin{figure}[t]
\includegraphics[width=0.45\textwidth]{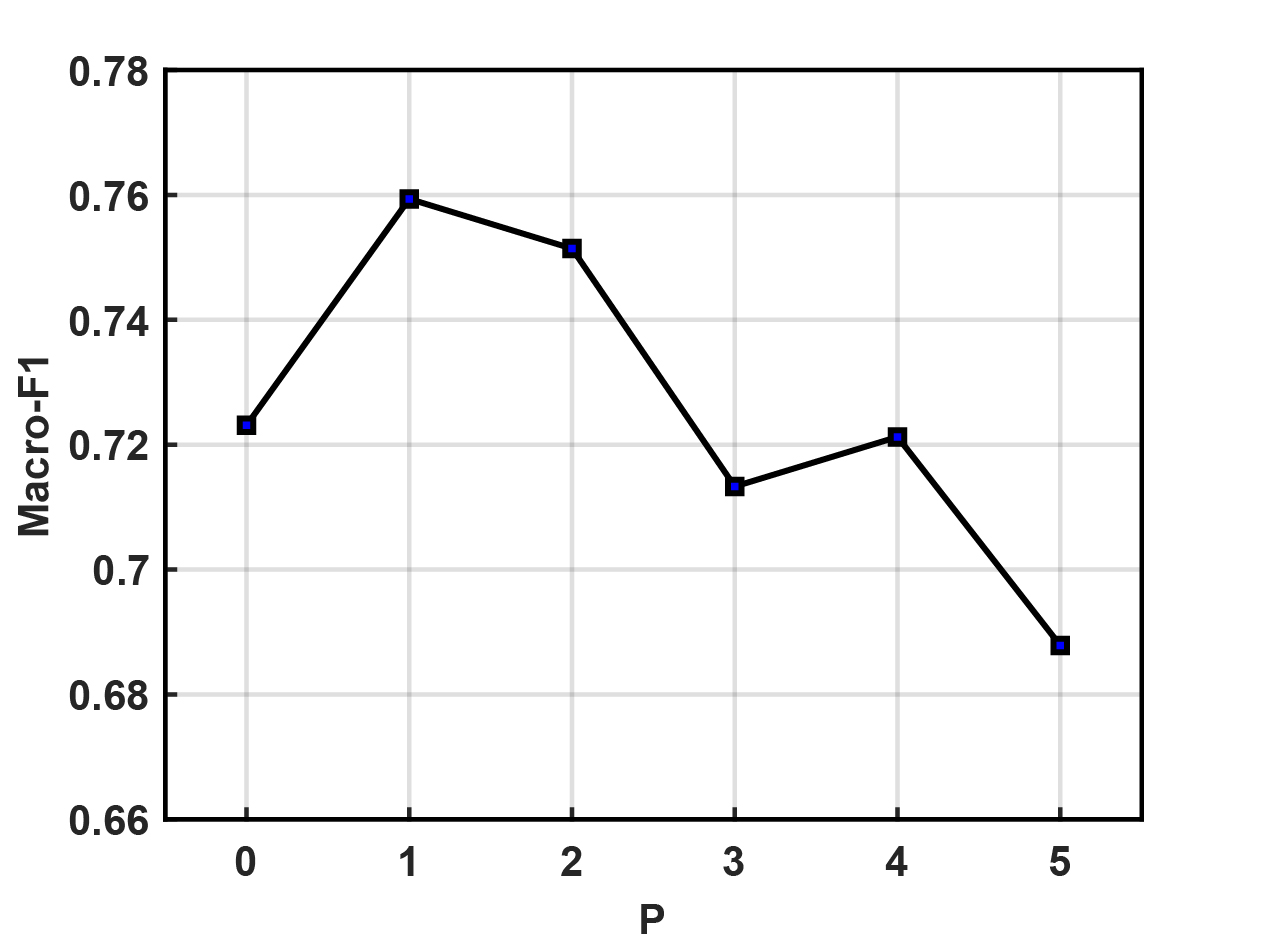}
\centering
\caption{The Macro-F1 scores of different \emph{P} values on \emph{Newsgroup} with $S^L$}
\label{Fig3}
\end{figure}

Reviewing Eq.\ref{Eq4}, we see that the parameter \emph{P} indicates how many pseudo-nearest neighboring categories will be used to compute the label membership degree prior, and $P=0$ implies that the prototype scheme is not applied.

By holding other parameters fixed, we examine the Macro-F1 scores of different values of \emph{P} over the set $\{0,1,2,3,4,5\}$ on \emph{Newsgroup} with $S^L$. As the results shown in Figure \ref{Fig3}, two observations are made. First, we can see that the Macro-F1 scores of $P=1,2$ are higher than that of $P=0$, indicating that the prototype scheme can effectively improve the classification performance. Second, the Macro-F1 scores become lower as the value of \emph{P} continues to increase. The reason is that although using more pseudo-nearest labels, \ie larger \emph{P} values, can effectively cover the relevant labels in $\alpha'$, it also introduces much more irrelevant labels, making $\alpha'$ even more noisy. Therefore, we suggest $P=1$ as the default setting.

\section{Conclusion}
\label{6}

In this paper, we develop a novel \baby method for dataless text classification with seed words. The main idea of \baby is to formulate a document-specific category prior, which can enrich and modify the supervision signals by self-exploring the available data and seed words. First, a novel prototype scheme is presented to better capture the label membership degree prior of documents, and second, the label frequency prior is estimated by seed word occurrences to further enrich the supervision. We conduct a number of experiments to evaluate the effectiveness of \baby. The experimental results indicate that \baby outperforms the traditional dataless methods, and it performs well with very few seed words. Additionally, \baby is even on a par with supervised methods in some settings. In the future, we plan to extend \baby to multi-label classification tasks.

\section*{Acknowledgment}

This research was supported the National Natural Science Foundation of China (NSFC) [No.61876071].

\label{}




\bibliographystyle{elsarticle-harv}
\bibliography{WSPTM}

\section*{Appendix}

In this Appendix, we derive the key update equations of \baby.

\vspace{0.5ex}
\noindent{\textbf{The posterior of topic assignment of each word token, \ie Eqs.\ref{Eq10} and \ref{Eq11}}}: 
By holding $\{\theta,\widehat \theta, \phi, \widehat \phi\}$ fixed, for each word token $w_{dn}$, the joint distributions of the word token and topic assignments are given as follows:
\begin{equation}
\label{EqA-1}
p\left(w_{dn}, c_{dn}=1, \, z_{dn}=k\right) = \theta_d^T \gamma_{w_{dn}} \, \theta_{dk} \phi_{kw_{dn}}, 
\end{equation}
\begin{equation}
\label{A-2}
p\left(w_{dn},c_{dn}=0, \, \widehat z_{dn}=g\right) = \left(1-\theta_d^T\gamma_{w_{dn}}\right) \widehat \theta_{dg} \widehat \phi_{gw_{dn}} 
\end{equation}
The marginal probabilities are given as follows: 
\begin{equation}
\label{EqA-3}
p\left(w_{dn}, c_{dn}=1\right) = \theta_d^T\gamma_{w_{dn}} \sum \nolimits_{i=1}^K\theta_{di}\phi_{iw_{dn}}, 
\end{equation}
\begin{equation}
\label{A-4}
p\left(w_{dn},c_{dn}=0\right) = \left(1-\theta_d^T\gamma_{w_{dn}}\right) \sum \nolimits_{i=1}^G\widehat \theta_{di}\widehat \phi_{iw_{dn}} 
\end{equation}
With the above formulas, we can estimate the posteriors of topic assignments by applying the Bayes rule:
\begin{equation}
\label{EqA-5}
p\left(z_{dn}=k\right) = \theta_d^T\gamma_{w_{dn}} \frac{\theta_{dk}\phi_{kw_{dn}}}{\sum \nolimits_{i=1}^K \theta_{di}\phi_{iw_{dn}}} \buildrel \Delta \over = N_{dnk},
\end{equation}
\begin{equation}
\label{EqA-6}
p\left(\widehat z_{dn}=g\right) = \left(1-\theta_d^T\gamma_{w_{dn}}\right) \frac{\widehat \theta_{dg}\widehat \phi_{gw_{dn}}}{\sum \nolimits_{i=1}^G \widehat \theta_{di}\phi_{iw_{dn}}} \buildrel \Delta \over = \widehat N_{dng}
\end{equation}
We kindly emphasize that we omit the notations $c_{dn}=1$ and $c_{dn}=0$ in Eqs.\ref{EqA-5} and \ref{EqA-6} to make equations simple.

\vspace{1.5ex}
\noindent{\textbf{The update equations of $\{\widehat \theta, \phi, \widehat \phi\}$, \ie Eqs.\ref{Eq121314}, 13, 14, and the ``initialization'' equation of $\theta$, \ie Eq.\ref{Eq15}}}: 
Given the current $\{N_{dnk}\}_{k=1}^K, \{\widehat N_{dng}\}_{g=1}^G$ of all word tokens and pre-computed word weights $\pi$, we can use them to generate soft occurrence numbers of distribution-specific samples. We take $\widehat \theta$ as an example. For each document $d$, we can regard $\{\sum \nolimits_{n=1}^{N_d}\pi(w_{dn})\widehat N_{dng}\}_{g=1}^G$ as the soft occurrence numbers of background topics, while we known that $\widehat \theta$ is drawn from a Dirichlet prior $\widehat \alpha$. Accordingly, we can regard this as a Dirichlet-Multinomial estimation, directly giving the update equation of $\widehat \theta$ as follows:
\begin{equation}
\label{EqA-7}
\widehat \theta_{dg} = \frac{\sum \nolimits_{n=1}^{N_d}\pi(w_{dn})\widehat N_{dng}+\widehat \alpha}{\sum \nolimits _{i=1}^G \sum \nolimits_{n=1}^{N_d}\pi(w_{dn})\widehat N_{dni}+ G\widehat \alpha} 
\end{equation}
The formulas of $\{\theta, \phi, \widehat \phi\}$ are similar to that of $\widehat \theta$, so we omit details.







\end{document}